\documentclass{amia}
\usepackage{lipsum} 

\usepackage{listings}
\usepackage{color}
\usepackage{makecell}
\usepackage{multirow}
\usepackage{float}
\usepackage{comment}
\usepackage{booktabs}
\usepackage{tabularx}
\usepackage{graphicx}
\usepackage[colorlinks=true]{hyperref}
\usepackage{caption}
\usepackage{subcaption}
\usepackage{paralist}
\usepackage{xspace}
\usepackage{wrapfig}

\newcommand{\sllm}{Llama 3.1-8B\xspace}

\begin{document}

\title{Enhancing Disease Detection in Radiology Reports Through Fine-tuning Lightweight LLM on Weak Labels}

\author{Yishu Wei, PhD$^1$, Xindi Wang, PhD$^2$, Hanley Ong, M.D.$^2$, Yiliang Zhou, MS$^1$, Adam Flanders, M.D.$^3$, George Shih, M.D.$^2$, Yifan Peng, PhD$^1$ }

\institutes{
    $^1$ Department of Population Health Sciences, Weill Cornell Medicine, New York;
    $^2$ Department of Radiology, Weill Cornell Medicine, New York;
    $^3$ Department of Radiology, Thomas Jefferson University Hospital, Philadelphia
}

\maketitle

\section*{Abstract}

\textit{Despite significant progress in applying large language models (LLMs) to the medical domain, several limitations still prevent them from practical applications. Among these are the constraints on model size and the lack of cohort-specific labeled datasets. In this work, we investigated the potential of improving a lightweight LLM, such as \sllm, through fine-tuning with datasets using synthetic labels. Two tasks are jointly trained by combining their respective instruction datasets. When the quality of the task-specific synthetic labels is relatively high (e.g., generated by GPT4-o), \sllm achieves satisfactory performance on the open-ended disease detection task, with a micro F1 score of 0.91. Conversely, when the quality of the task-relevant synthetic labels is relatively low (e.g., from the MIMIC-CXR dataset), fine-tuned \sllm is able to surpass its noisy teacher labels (micro F1 score of 0.67 v.s. 0.63) when calibrated against curated labels, indicating the strong inherent underlying capability of the model. These findings demonstrate the potential of fine-tuning LLMs with synthetic labels, offering a promising direction for future research on LLM specialization in the medical domain.}

\section{Introduction}

There have been extensive studies on applying large language models (LLMs) in the medical domain. However, several challenges must be overcome before their practical application is feasible. From the model perspective, privacy concerns limit the usage of commercial LLMs, such as GPT-4, since patient data involves highly sensitive information, and current de-identification techniques do not completely protect privacy \cite{rothstein2010deidentification}. 
Additionally, the financial, computational, and technical demands of deploying super-large, strong LLMs (e.g., Llama 3.1-405B) pose a significant challenge for hospitals. 
Consequently, lightweight LLMs (e.g., \sllm) are considered more feasible, though they often sacrifice the performance significantly \cite{zhang2024closing}. 
From the data perspective, although there is an abundance of public datasets, they are often disease-specific and do not reflect the diversity found in hospital patient cohorts. Additionally, while hospitals possess vast amounts of patient data, the available labels are often of poor quality or entirely missing.

Addressing these challenges can be achieved by fine-tuning lightweight LLMs using synthetic or weakly labeled data. This strategy draws from traditional deep learning distillation techniques, where predictions from a strong model are used to ``teach" a less powerful model. For instance, Pangakis et al. \cite{pangakis2024knowledge} have demonstrated that fine-tuning on LLM-generated data can yield outcomes comparable to human-annotated data. While most previous research \cite{huang2024leveraging, tang2023does, kumichev2024medsyn} has shown the effectiveness of this approach mainly when the baseline model performs modestly, few studies indicate that substantial improvements can still be achieved in the radiology domain \cite{gu2024chex}.



To bridge this gap, this study aims to fine-tune \sllm using weak labels on two radiology-specific tasks. The first task involves multiple-choice lung disease classification on radiology reports, where the model predicts diseases from a predetermined list. For this, a rule-based labeler, Negbio\cite{peng2018negbio}, was used to extract 13 labels from radiology reports and then construct instructions for fine-tuning lightweight LLMs. Despite the inherent noisiness of these labels, we found the fine-tuned \sllm, demonstrated notable performance improvements on human-curated labels, even exceeding the accuracy of the Negbio. This result is somewhat unusual in traditional deep learning models, where the student model generally underperforms compared to the teacher model.

The second task focuses on open-ended lung disease detection, requiring the LLM to extract abnormal findings from radiology reports corresponding to ICD-10 codes. For this task, the lightweight \sllm was fine-tuned using synthetic labels generated by GPT-4o. As illustrated in previous studies by Liu et al. \cite{liu2023exploring}, GPT-4o has demonstrated robust performance across various medical tasks. We observed that the fine-tuned \sllm achieved performance metrics nearly on par with GPT-4o. 

To summarize, our contributions are as follows.
\begin{inparaenum}[(1)]
    \item We demonstrated the strong potential of fine-tuning a lightweight LLM specialized in the radiology domain through instruction-based learning with weak labels.
    \item We explored jointly fine-tuning a single LLM on two distinct tasks by mixing instruction sets derived from different datasets. This mixed dataset approach could optimize the performance of the fine-tuned LLMs across both tasks.
\end{inparaenum}

\section{Materials and Methods}\label{methods}

This study proposes a multi-task learning framework designed to enhance disease detection in radiology reports by leveraging both structured and unstructured tasks (Figure~\ref{fig:example}). Our approach adopts instruction tuning for a multiple-choice disease classification task and an open-ended disease detection task. 

\begin{figure}
    \centering
    \begin{subfigure}[b]{\textwidth}
\footnotesize
\begin{tabularx}{\textwidth}{X}
\toprule
\textbf{Prompt}: \\
You are given a clinical report. Your task is to identify the diseases that the patient have and list them. The result list should be formatted exactly as [disease1, disease2, ...] and be on a single line. \\
\\
Only use diseases from this list:\\
(`Atelectasis', `Cardiomegaly', `Consolidation', `Edema', `Enlarged Cardiomediastinum', `Fracture', `Lung Lesion', `Lung Opacity', `Pleural Effusion', `Pleural Other', `Pneumonia', `Pneumothorax', `Support Devices')\\
\\
Guidelines:\\
** Be careful about negation. E.g., do not include Pneumonia if the text says "no Pneumonia"\\
** Only include diseases that are certain. Entities that are 'likely', 'possibly' should not be returned. \\
** After output the list, provide the evidence in the text for each disease.    \\
\\
\textbf{Report to Analyze}:\\
Findings \\
Tubes and Lines: None \\
Lungs: There are bilateral patchy pulmonary opacities \\
Pleura: Small pleural effusions noted. \\
Heart/Mediastinum: The cardiac silhouette is mildly enlarged. The mediastinal silhouette is within normal limits. \\
Bones: There are degenerative changes of osseous structures. \\

Impression \\
Patchy bilateral pulmonary opacities likely represent pneumonia \\
\\
\textbf{Output}: \\
The diseases are: [`Lung Opacity', `Pleural Effusion', `Cardiomegaly']\\
\bottomrule
\end{tabularx}
    \caption{Multiple-choice disease classification}
    \vspace*{2em}
    \end{subfigure}
    \begin{subfigure}[b]{\textwidth}
\footnotesize

\begin{tabularx}{\textwidth}{X}
\toprule
\textbf{Prompt}: \\
You are a radiologist. Your task is to do entity recognition, which will extract phrases in the report that represent a potential ICD-10 code.
Your result should be a list, i.e., [phrase 1, phrase 2, ...]. The phrase should fully contain the indication of an ICD-10 code, but also should be as short as possible.\\
\\
Guidelines:\\
** Extract phrases that directly correlate to an ICD-10 code.\\
** Keep the phrases as concise as possible while retaining their full meaning.\\
** Provide the reasoning behind each extracted phrase, including the corresponding ICD-10 code.\\
** Include only the conditions that the patient has. For example, if the report states `No pneumothorax,' do not include pneumothorax.\\
** If there are no conditions or findings that correlate to an ICD-10 code, directly output `Output: []'\\
\\
Example Reports and Outputs:\\
1. \{example1\}\\
2. \{example2\}\\
3. \{example3\}\\
\\
\textbf{Report to Analyze}:\\
Findings \\
Tubes and Lines: None \\
Lungs: There are bilateral patchy pulmonary opacities \\
Pleura: Small pleural effusions noted. \\
Heart/Mediastinum: The cardiac silhouette is mildly enlarged. The mediastinal silhouette is within normal limits. \\
Bones: There are degenerative changes of osseous structures. \\

Impression \\
Patchy bilateral pulmonary opacities likely represent pneumonia \\
\\
\textbf{Output}:\\
`Patchy pulmonary opacities', `Small pleural effusions', `Mildly enlarged cardiac silhouette', `Degenerative changes of osseous structures'\\
\bottomrule
\end{tabularx}

\caption{Open-ended disease detection}
\end{subfigure}
\caption{Samples of instruction, input, and output for disease detection in radiology reports.} 
\label{fig:example}
\vspace{1em}
\end{figure}

\subsection{Problem Formulation}

\paragraph{Multiple-choice Disease Classification}

The first task in our multi-task learning framework is a multiple-choice disease classification problem, aimed at identifying a specific lung disease or abnormal findings from a radiology report (Figure~\ref{fig:example}a). In this setup, the input consists of a natural language question representing the radiology report and a list of candidate answers representing potential diagnoses. 
Here, we focus on 13 common lung diseases: Atelectasis, Cardiomegaly, Consolidation, Edema, Enlarged Cardiomediastinum, Fracture, Lung Lesion, Lung Opacity, Pleural Effusion, Pleural Other, Pneumonia, Pneumothorax, Support Devices. A sample answer for this problem can be \texttt{[`Fracture', `Lung Opacity', ...]}.

\paragraph{Open-ended Disease Detection}

The second task is an open-ended disease detection problem, aimed at enabling the model to generate a free-form response from a radiology report (Figure \ref{fig:example}b). Unlike the multiple-choice task, this task doesn't restrict the model to a predefined list of potential lung diseases or abnormal findings. Instead, it prompts the model to identify and describe diseases mentioned in the radiology report. Therefore, open-ended disease detection allows for a deeper evaluation of the LLM's comprehension, contextual understanding, and ability to enumerate various abnormal findings. A sample answer for this problem can be \texttt{[`Small pleural effusions', `Mildly enlarged cardiac silhouette', ...]}, with these terms being directly extracted from the original report.

\subsection{Data Preparation}

To support the two tasks, we utilized three chest X-ray datasets. For the multiple-choice disease classification task, we employed the MIMIC-CXR dataset \cite{Johnson2019-dz}. For the open-ended disease detection task, we used the NIH-CXR/MIRDC \cite{zhou2024evaluating} and WCM \cite{pandey2020extraction} datasets. Each dataset comprises chest X-ray images and their corresponding radiology reports, which were processed to extract the relevant information necessary for model training. The detailed statistics of these datasets are summarized in Table~\ref{tab:data_summary}.
\begin{table}
\caption{Statistics of the dataset.}
\label{tab:data_summary}
\vspace{-1em}
\centering
\begin{tabular}{lllr} 
\toprule
Dataset & Usage & Annotator & Reports \\ 
\midrule
Multiple choice disease detection\\
\hspace{2em}MIMIC-CXR & Instruction-based fine-tuning& NegBio  & 4,500\\
\hspace{2em}Curated MIMIC-CXR & Evaluation & Radiologists & 100 \\ 
Open-ended disease detection\\
\hspace{2em}WCM & Instruction-based fine-tuning & GPT4-o& 9,000\\
\hspace{2em}NIH-CXR/MIRDC & Prompting; Evaluation & Radiologists & 100\\
\bottomrule
\end{tabular}
\end{table}

\paragraph{NIH-CXR/MIRDC} This dataset consists of 100 radiology reports independently annotated by a cohort of radiologists, including two attending physicians and three residents \cite{zhou2024evaluating}. Fifty of the reports were randomly selected from the NIH-CXR dataset, with corresponding reports dictated by one attending physician and three radiology residents. The remaining 50 chest radiographs, along with de-identified free-text radiology reports, were randomly selected from the Medical Imaging and Data Resource Center (MIDRC) \cite{midrc2022-uz}. 

\paragraph{WCM} This dataset consists of 9,000 thoracoabdominal chest X-ray reports from heart failure patients treated at New York-Presbyterian/Weill Cornell Medical Center. These patients were admitted and discharged with ICD-9 Code 428 or ICD-10 Code I50 between January 2008 and July 2018 \cite{pandey2020extraction}. For each report, only the findings and impression sections were used. The data was reviewed and de-identified in accordance with institutional review board (IRB) protocols.

\paragraph{MIMIC-CXR} This dataset is a large collection of chest X-rays sourced from Beth Israel Deaconess Medical Center, covering the period between 2011 and 2016 \cite{Johnson2019-dz}. For computational efficiency, 100,000 reports containing either a ``Findings'' or ``Impression'' section were randomly selected. For this study, two sets of labels were derived. The first set was generated using NegBio, an open-source, rule-based labeling tool \cite{peng2018negbio}, which extracted labels for thirteen diseases. 
The second set consists of 100 randomly selected reports from MIMIC-CXR and was manually annotated by one radiologist (HO) and one domain expert (YW). To assess inter-rater agreement, the Cohen's Kappa coefficient \cite{waisberg2023gpt} was utilized. The coefficient was 0.53 between the MIMIC labels and HO, 0.52 between the MIMIC labels and YW, and notably higher at 0.88 between HO and YW.
Two annotators discussed the disagreement and a consensus was reached.
The final labels were used as the ground truth for this study.

\subsection{Approaches}

\begin{figure}[H]
    \centering
\includegraphics[width=.6\textwidth]{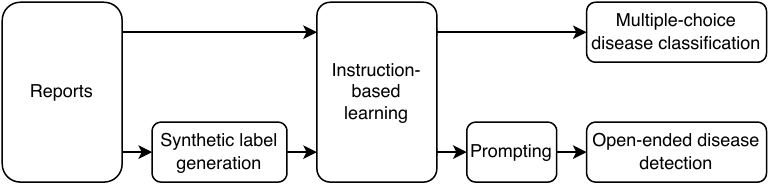}
    \caption{Overview of the experiment setup} 
    \label{fig:overview}
    \vspace{1em}
\end{figure}

In our study, we employed \sllm as the foundational model. It functions as a decoder-only model, processing inputs and generating text autoregressively. 
We fine-tuned \sllm jointly on two distinct tasks by mixing instruction sets derived for both tasks (Figure \ref{fig:overview}). This joint approach was utilized to optimize the performance of the \sllm model across both tasks. 

\paragraph{Multiple-choice Disease Classification}

This task was conducted on the MIMIC-CXR dataset. 
In the instruction-based learning approach, we employ labels generated by NegBio as outputs. The instructions for fine-tuning are relatively simplified, lacking the Guidelines and Examples, in contrast to the more detailed prompts used in the final detection task.

After the fine-tuning process, \sllm was provided with the 13 diseases and prompted to predict diseases from radiology reports in a zero-shot fashion. Details of the prompts are provided in Figure \ref{fig:example}a.



\paragraph{Open-ended Disease Detection}

In the open-ended disease detection task, \sllm was prompted to extract phrases from the radiology report that can be mapped to ICD-10 codes in a few-shot manner. 

In the absence of predefined labels, we implemented a two-step approach. Initially, labels were synthesized using GPT-4o (gpt-4o-2024-05-13, \url{https://openai.com/index/hello-gpt-4o/}). 
Specifically, 9,000 samples were selected from the WCM dataset. 
GPT-4o was then prompted to ``\textit{extract phrases in the report
that represents a potential ICD-10 code}" (Table \ref{tab:prompt_generation}). To improve the accuracy of the generated labels, seed examples were also incorporated in the prompt. In particular, eleven reports were selected from the NIH-CXR/MIRDC dataset to serve as seed examples. For each label generation, three out of these eleven seed examples were randomly chosen and included in the prompt.


Following this, instruction-based fine-tuning was carried out alongside the earlier task. To enhance performance, and in contrast to the multiple-choice disease classification task which was approached in a zero-shot fashion, three examples from the NIH/MIRDC dataset were provided in the final disease detection prompt (Figure \ref{fig:example}b). 
%
\begin{table}
\caption{Prompt used for synthetic label generation}
\label{tab:prompt_generation}
\vspace{-1em}
\centering
\begin{tabularx}{\textwidth}{X}
\toprule
    You are a radiologist. Your tasks is to do disease detection, which will extract phrases in the report that represents a potential ICD-10 code.
    Your result should be a list, i.e. [phrase 1, phrase 2, ...]. The phrase should fully contain the indication of a ICD-10 code, but also should be as short as possible.\\
\\
    Guidelines:\\
    ** Extract phrases that directly correlate to an ICD-10 code.\\
    ** Keep the phrases as concise as possible while retaining their full meaning.\\
    ** Provide the reasoning behind each extracted phrase, including the corresponding ICD-10 code.\\
    ** Include only the conditions that the patient has. For example, if the report states 'No pneumothorax,' do not include pneumothorax.\\
    ** If there are no conditions or findings that correlate to an ICD-10 code, directly output 'Output: []'\\
\\
Example Reports and Outputs: \\
\bottomrule
\end{tabularx}
\end{table}

\subsection{Implementation Details and Evaluations}

We fine-tune \sllm using the Low-Rank Adaptation (LoRA) technique \cite{hu2021lora} for both tasks. LoRA was specifically applied to the attention modules, targeting the q-proj and v-proj matrices. The LoRA rank is set to 8, with a scaling factor (LoRA alpha) of 16. The AdamW optimization algorithm was employed, utilizing a learning rate of $3 \times 10^{-4}$ and a weight decay of 0.01. 
The instruction datasets from both tasks—the open-ended and multiple-choice—were merged and shuffled for fine-tuning.

The training was conducted on a single NVIDIA A100 GPU, with all results reported at a consistent temperature setting of 0.1 across various configurations.

To evaluate the performance of fine-tuned LLM on multiple-choice disease classification tasks, we reported micro-averaging precision, recall, and F1 across all diseases. Since the responses are selected from a predefined list of possible diseases, a generated disease is deemed correct only if it exactly aligns with an option from that list. 

To evaluate the effectiveness of open-ended disease detection, we reported the micro-averaging accuracy, recall, and F1 score metrics. 
These metrics consider a prediction correct (true positive) if two consecutive words match between the ground truth and the generated phrases. 
For instance, phrases such as `bilateral peribronchial cuffing' and `peribronchial cuffing' are considered as a match. 
Several reasons underpin the decision to use this matching criteria rather than requiring an exact match. 
First, most medical terms contain fewer than two words, and GPT-4o is prompted to generate concise phrases.
Second, labeling by radiologists may vary.
And thirdly, unigram can be easily misleading and informative, particularly with function words like ``of" or ``for". This matching criterion is also supported by a detailed error analysis, affirming its validity.

\section{Results and Discussion}

\subsection{Findings from Multiple-choice Disease Classification}
\label{mq_result}

\paragraph{Effectiveness of Fine-tuning}

We fine-tuned the \sllm model using the NegBio-generated labels and evaluated its performance against human-curated labels. 
Table \ref{tab:joint_mimic} shows that fine-tuning substantially improved the model’s performance (0.67 vs 0.54 on human-curated labels). 
These results indicate that instruction-based fine-tuning can significantly bolster the model’s classification capabilities, even when benchmarked against more sophisticated, curated label sets in the radiology domain. 



\begin{table}
\caption{Performance of models on the multiple-choice disease classification task using micro-precision, micro-recall, and micro-F1 scores. The model was trained on the combination of WCM and MIMIC-CXR datasets and then tested on human-curated data from the MIMIC-CXR dataset.}
\label{tab:joint_mimic}
\vspace{-1em}
\centering
\begin{tabular}{lrrrrrr} \toprule
& \multicolumn{3}{c}{Fine-tuning Data} &\\ 
\cmidrule(rl){2-4}
Model & Total & WCM & MIMIC & Precision& Recall& F1\\ \midrule
\sllm
 & 0 & 0 & 0 &0.46&0.65&0.54\\ 
 & 1,500 & 1,000 & 500&0.44&0.62&0.51\\
 & 3,000 & 2,000 & 1,000&0.45&0.64&0.53\\ 
 & 4,500 & 3,000 & 1,500&0.53&0.67&0.59\\ 
 & 7,500 & 5,000 & 2,500&0.61&0.72&0.66\\ 
 & 9,000 & 6,000 & 3,000& 0.62& 0.72&0.67\\
 & 13,500 & 9,000 & 4,500& 0.69&0.59&0.64\\ 
NegBio &-&-&-&0.82&0.51&0.63\\ 
GPT4-o&-&-&-&0.83&0.89&0.86\\
 \bottomrule
\end{tabular}
\end{table}



Significantly, when the fine-tuning utilizes 9,000 samples (3,000 from MIMIC-CXR and 6,000 from WCM), \sllm achieved a recall of 0.72, substantially better than NegBio, which achieved a recall of 0.51. This performance indicates that with an optimal level of fine-tuning, the model’s intrinsic qualities enable it to generalize more effectively and provide more accurate disease classifications. 

Additionally, we observed that the fine-tuned \sllm has demonstrated the ability to surpass the performance of the NegBio, achieving an F1 score of 0.67 compared to 0.63. In this study, \sllm is substantially larger and technologically superior compared to the NegBio. This phenomenon is noteworthy as it suggests that under certain conditions, the super intelligent `student' model can indeed exceed the weak intelligent `teacher' model’s limitations. 

\paragraph{Effectiveness of training data size and learning rate}

Our findings suggest that there exists an optimal fine-tuning intensity, which is characterized by the amount of data used and/or can be adjusted through learning rate modifications. 

As for the amount of fine-tuning data, Table \ref{tab:joint_mimic} suggests that at fine-tuning intensities of 9,000 samples, the model effectively harnesses both its intrinsic capabilities and the specific data from the NegBio-labeled dataset, resulting in the best performance outcomes. 
Subsequently, the model began to \textbf{overfit} the NegBio-generated labels, which unfortunately resulted in a degradation of performance when evaluated against the human-curated labels. 

\begin{wrapfigure}{R}{.45\textwidth}
    \centering
    \includegraphics[width=\linewidth]{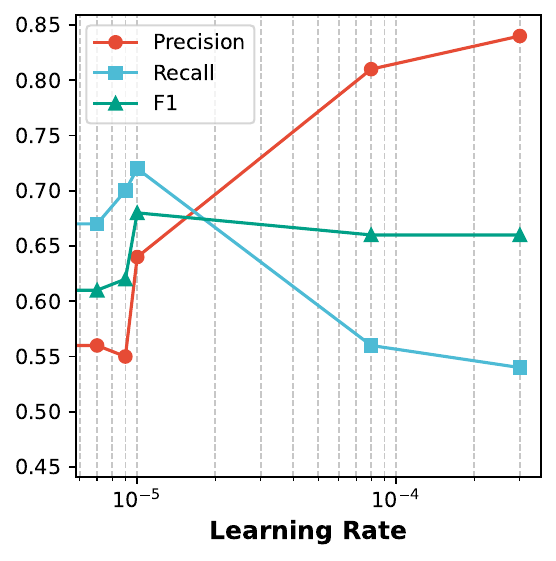}
    \caption{Performance of \sllm on the multiple-choice disease classification task with different learning rates. The model was trained on 100,000 samples from the MIMIC-CXR dataset and then tested on human-curated data from the same MIMIC-CXR source.}
    \label{fig:mimic_yishu_lr}
\end{wrapfigure}
As for the learning rates, we evaluated the performance of \sllm with different learning rates. The model was trained on 100,000 samples from the MIMIC-CXR dataset and then tested on human-curated data from the same MIMIC-CXR source.
Fig \ref{fig:mimic_yishu_lr} achieved the best performance (F1 of 0.68) when the learning rate is $1\times 10^{-5}$. These findings indicate that careful hyperparameter tuning is critical to achieving optimal performance.

\subsection{Results on open-ended disease detection}
\paragraph{Effectiveness of Fine-tuning}

Table \ref{tab:joint_open} summarizes the performance of models on the open-ended disease detection task. The model was trained on the combination of WCM and MIMIC-CXR datasets and then tested on the NIH/MIDRC dataset. 
Notably, as the volume of training data increases, the performance of \sllm (F1 of 0.91) approached that of the larger GPT-4o (F1 of 0.93). This demonstrates the effectiveness of instruction-based fine-tuning in enabling a smaller model to reach performance levels comparable to those of a larger one.

Furthermore, this study distinguishes itself from previous research by documenting performance improvements at an already high baseline. Initially, even before the fine-tuning process, \sllm demonstrated a strong F1 score of 0.83, which was enhanced to 0.91 through fine-tuning. This improvement emphasizes the potential of fine-tuning strategies, not only to elevate model performance but also to maintain high-performance standards even from a higher starting point.

Like the multiple-choice disease classification task, continuously adding fine-tuning data does not necessarily enhance the final outcome in open-ended disease detection. Specifically, there was no improvement in performance when the sample size was increased to 13,500. is suggests that beyond a certain point, additional data and prolonged training does not contribute to, and may even hinder, the efficiency and effectiveness of the model's learning process, indicating a potential saturation point where the model has maximized its learning from the available data.

\begin{table}
\caption{Performance of models on the open-ended disease detection task using micro-accuracy, micro-recall, and micro-F1 scores. The model was trained on the combination of WCM and MIMIC-CXR datasets and then tested on the NIH/MIDRC dataset.}
\label{tab:joint_open}
\centering
\vspace{-1em}
\begin{tabular}{lrrrrrr} 
\toprule
& \multicolumn{3}{c}{Fine-tuning Data} &\\ 
\cmidrule(rl){2-4}
Model & Total & WCM & MIMIC & Precision & Recall& F1\\ \midrule
\sllm
 & 0 & 0 & 0 &0.86&0.81&0.83 \\
 & 1,500 & 1,000 & 500 & 0.82& 0.90& 0.86\\
 & 3,000 & 2,000 & 1,000 & 0.82&0.89&0.85\\
 & 4,500 & 3,000 & 1,500 & 0.88& 0.90& 0.89\\
 & 9,000 & 6,000 & 3,000&0.92&0.91&0.91\\ 
 & 13,500 & 9,000 & 4,500& 0.93&0.86&0.89\\
GPT4-o & - & - & - & 0.92 & 0.95 & 0.93 \\
\bottomrule 
\end{tabular}
\end{table}

\paragraph{Error analysis}

We conducted an error analysis to categorize the error types. Here we used the model with the best performance (fine-tuned on 9,000 samples).

First, we observed that paraphrasing of the ground truth occurred ten times (Table \ref{tab:inconsistency}). When annotating labels, radiologists may paraphrase the original text from the reports, whereas the LLM is instructed to directly extract original phrases. Table \ref{tab:paraphrase} demonstrates some examples. For example, a phrase like ``emphysematous and fibrotic changes of the lungs" might be separated by the radiologist into ``emphysematous changes" and ``fibrotic changes.'' Secondly, common misspellings were noted, such as ``atelectasis'' being recorded as ``atalectasis'' in the radiology reports. Finally, we observed the discrepancies between machine-generated and radiologist-generated descriptions. For instance, the phrase ``left PICC line in superior vena cava" is captured as ``left PICC" by the LLM, while the radiologist captures it as ``PICC line in superior vena cava".

Second, despite being prompted otherwise, \sllm occasionally included terms, such as ``likely" or ``possibly" 18 times, which accounted for the majority of errors. 

Third, several lung diseases were frequently missed by \sllm, such as ``diffuse/patchy pulmonary opacities" and ``mild degenerative changes of osseous structures". This indicates a potential area for improvement in how the model detects and interprets specific types of pulmonary conditions.

Finally, it's worth noting that some cases missed by GPT-4 might suggest inherent uncertainty in the ground truth itself. For instance, ``Pulmonary opacities" was mentioned seven times, making up 33.3\% of the total total error, excluding paraphrased and uncertain predictions. Upon further analysis, GPT-4o explained that \textit{``diffuse pulmonary opacities" does not directly correspond to a specific ICD-10 code as it is more of a descriptive term used in radiology.} Indeed, it may be difficult to assign an ICD-10 code to this particular term, as it is a broad radiologic descriptor that could represent a wide range of potential diagnoses depending on clinical context and concurrent presence of other radiologic findings. Since the \sllm model was fine-tuned using labels generated by GPT-4o, it reflects the interpretive perspectives of the teacher model.


\begin{table}
\caption{Error analysis in the open-ended disease detection}
\label{tab:inconsistency}
\centering
\vspace{-1em}
\begin{tabularx}{0.8\textwidth}{lXr} \toprule
Error type & Example & Count \\\midrule
False positive 
& Paraphrase, See Table \ref{tab:paraphrase} & 10\\
& Uncertain: ``possibly", ``likely", ...& 18\\
& cervical spinal fusion hardware & 1 \\
\midrule
False negative 
& diffuse/patchy pulmonary opacities & 7\\
& mild degenerative changes of osseous structures &5 \\ 
& feeding tube enters the stomach &3 \\ 
& cardiac silhouette upper limit of normal size & 1\\ 
& consistent with history of mesothelioma & 1\\ 
& findings related to prior cabg &1 \\ 
& low lung volume  & 1\\ 
& obscuring of the right paratracheal line  & 1\\ 
& retrocardiac opacity  &1 \\ 
& unchanged elevation of the right hemidiaphragm & 1 \\ 
\bottomrule
\end{tabularx}
\vspace{2em}
\end{table}

\begin{table}
\caption{Paraphrase of ground truth appeared in the open-ended disease detection.}
\label{tab:paraphrase}
\centering
\vspace{-1em}
\begin{tabularx}{0.8\linewidth}{>{\raggedright\arraybackslash}p{15em}>{\raggedright\arraybackslash}X>{\raggedright\arraybackslash}X}
\toprule
Reason & LLM-generated label & Ground truth \\ \midrule
Misspelling & atalectasis & atelectasis  \\ \midrule
Discrepancies between machine and radiologist-generated descriptions & left PICC & PICC line in superior vena cava  \\\midrule 
Conjunction & emphysematous and fibrotic changes of the lungs  & emphysematous changes, fibrotic changes \\ \cmidrule{2-3}
 & hyperexpanded and hyperlucent lungs & hyperexpanded lungs,  hyperlucent lungs \\
\bottomrule
\end{tabularx}
\end{table}

\subsection{Evaluating joint vs separate fine-tuning of models}

\begin{figure}
    \centering
    \includegraphics[width=.9\linewidth]{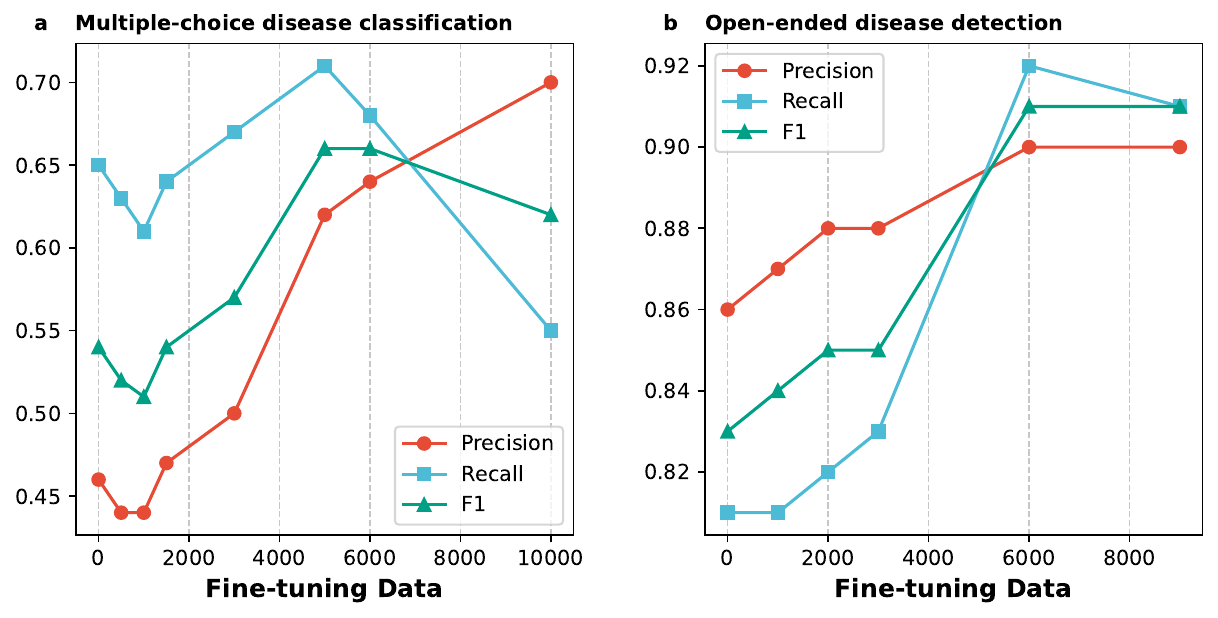}
    \caption{Performance of \sllm using micro-precision, micro-recall, and micro-F1 scores. (a) Multiple-choice disease classification. The model was trained on the MIMIC-CXR datasets \textbf{only} and then tested on human-curated data from the MIMIC-CXR dataset. (b) Open-ended disease detection. The model was trained on the WCM dataset \textbf{only} and then tested on the NIH/MIDRC dataset. 
    Learning rate was $3\times 10^{-4}$.}
    \label{fig:separate_mimic}
\end{figure}

Additionally, we conducted experiments to fine-tune the model separately for each task.
For the multiple-choice task, as the instruction dataset sample size was increased to 5,000, the peak F1 score of 0.66 was achieved (Figure~\ref{fig:separate_mimic}a). However, as the sample size continued to expand, reaching 10,000, there was a noticeable decline in the F1 score to 0.62. For the open-ended detection task (Figure~\ref{fig:separate_mimic}b), the highest performance was reached when the sample size increased to 6,000 (F1 score of 0.91). 
The performance outcomes of both joint and separate training methods proved to be similar across the tasks, showing no significant differences in observations.

\subsection{Limitations}\label{discussion}

Firstly, the open-ended disease detection task with radiology reports is relatively simple compared to the complexities inherent in real-world clinical disease diagnosis. Radiology reports, unlike unstructured clinical notes, typically adhere to a structured format, utilize consistent terminologies, and contain densely packed information, which simplifies the task of identifying mentions of diseases within the text. Despite these simplifications, the model's ability to approach the performance of human experts in an open-ended context is noteworthy. It highlights the potential of LLMs to evolve into versatile tools for medical assistance. Ultimately, this work can be seen as a foundational step towards developing more advanced models capable of managing the intricacies of open-ended disease diagnosis in real clinical settings.

Secondly, due to resource constraints with available radiologists, we managed to label only about 100 instances for testing. Obtaining a larger volume of annotated data would enable a more thorough evaluation of the performance and generalization capabilities of the LLMs. Expanding the dataset could significantly contribute to enhancing the reliability of our findings and provide deeper insights into the model's abilities across diverse clinical settings.

Lastly, this study did not engage in extensive prompt engineering beyond including a few examples for the model. Enhancements in recall could potentially be achieved by generating multiple outputs and employing ensembling techniques to integrate the results. Similarly, precision could be improved by implementing self-refinement or verification methods, which we did not explore in this research. Such strategies could further optimize the model's performance, potentially leading to more accurate and robust outcomes in disease detection tasks.

\section{Conclusion}\label{conclusion}

In this study, we explored the efficacy of fine-tuning a lightweight LLM using synthetic labels. We simultaneously fine-tuned the model for a multiple-choice disease classification task and an open-ended disease detection task through instruction-based learning. In the multiple-choice task, \sllm was capable of outperforming its rule-based teacher (i.e., NegBio), showcasing the model's intrinsic capabilities. In the open-ended detection task, GPT-4o delivered expert-level accuracy, which \sllm effectively adopted through instruction-based fine-tuning. These experiments, which span the spectrum from lower to higher synthetic label quality, collectively underscore the substantial potential of synthetic labels in boosting LLM performance across diverse tasks.

\subparagraph{Acknowledgments}
This work was supported by the National Science Foundation Faculty Early Career Development (CAREER) award number
2145640, the Intramural Research Program of the National Institutes of Health, and the Amazon Research Award. The Medical Imaging and Data Resource Center (MIDRC) is funded by the National Institute of Biomedical Imaging and Bioengineering (NIBIB) of the National Institutes of Health under contract 75N92020D00021 and through The Advanced Research Projects Agency for Health (ARPA-H).
We would like to thank Dr. Gongbo Zhang for his feedback.

\makeatletter
\renewcommand{\@biblabel}[1]{\hfill #1.}
\makeatother

\bibliographystyle{vancouver}
\bibliography{amia}

\end{document}